\documentclass[11pt]{article} 
\usepackage{rldmsubmit,palatino}
\usepackage{graphicx}
\usepackage[round]{natbib}  
\usepackage{color}
\usepackage{url}
\usepackage{graphicx}
\usepackage{amssymb}
\usepackage[utf8]{inputenc} 
\usepackage[T1]{fontenc}    
\usepackage{url}            
\usepackage{booktabs}       
\usepackage{amsfonts}       
\usepackage{nicefrac}       
\usepackage{microtype}      
\usepackage{amsmath}  		
\usepackage{amsfonts}		
\usepackage{algorithm}		
\usepackage{algorithmic}	
\usepackage{wrapfig}		
\usepackage{graphicx}		
\usepackage{subcaption}		
\usepackage{float} 			
\usepackage{graphicx}		
\usepackage{caption}		
\usepackage{dirtytalk}		
\usepackage[threshold=0]{csquotes}
\setcitestyle{aysep{ }}		
\usepackage{xcolor}			
\usepackage[margin=1in]{geometry}
\usepackage[hidelinks]{hyperref}
\usepackage{dsfont}
\usepackage[bottom]{footmisc}
\usepackage{amssymb}

\DeclareMathOperator{\E}{\mathbb{E}}

\title{Making Meaning: Semiotics Within\\ Predictive Knowledge Architectures}

\author{
Alex Kearney \\
Department of Computing Science\\
University of Alberta\\
Edmonton, Alberta, Canada \\
\texttt{kearney@ualberta.ca} \\
\And
Oliver Oxton \\
Department of Philosophy\\
University of Waterloo\\
Waterloo, Ontario, Canada \\
\texttt{opo.xton@gmail.com} \thanks{For additional discussion on the epistemology of predictive knowledge and how we can view predictive knowledge architectures as having knowledge, please refer to Kearney \& Pilarski, "When is a Prediction Knowledge?", accepted to RLDM 2019.}\\
}

\begin{document}

\maketitle

\begin{abstract}
Within Reinforcement Learning, there is a fledgling approach to conceptualizing the environment in terms of predictions. Central to this predictive approach is the assertion that it is possible to construct ontologies in terms of predictions about sensation, behaviour, and time---to categorize the world into entities which express all aspects of the world using only predictions. This construction of ontologies is integral to predictive approaches to machine knowledge where objects are described exclusively in terms of how they are perceived. In this paper, we ground the Pericean model of semiotics in terms of Reinforcement Learning Methods, describing Peirce's Three Categories in the notation of General Value Functions. Using the Peircean model of semiotics, we demonstrate that predictions alone are insufficient to construct an ontology; however, we identify predictions as being integral to the meaning-making process. Moreover, we discuss how predictive knowledge provides a particularly stable foundation for semiosis\textemdash the process of making meaning\textemdash and suggest a possible avenue of research to design algorithmic methods which construct semantics and meaning using predictions.
\end{abstract}




\acknowledgements{Thanks to Patrick Pilarski, Johannes G\"unther, Melissa Woghiren, Anna Koop, and Niko Yasui for feedback on early drafts of this manuscript. Thanks to Dylan Jones for insightful discussion. This work was supported in part by the Alberta Machine Intelligence Institute, Alberta Innovates, the Natural Sciences and Engineering Research Council, the Canada Research Chairs program, and by Borealis AI through their Global Fellowship Award.}

\startmain

\section{Predictive Approaches to Machine Knowledge}

A foundational problem of machine intelligence is being able to conceptualize the environment---to construct ontologies of categories and concepts which enable meaningful decision making and problem solving. Furthermore, it is normally assumed that such ontologies are dependent upon knowledge; and so it is no surprise that each approach to machine intelligence\textemdash from expert systems to machine learning methods\textemdash aims to build systems approaching human-level understanding of their environment based on differing assumptions about what knowledge is and what counts as it. Within Reinforcement learning, there is a fledgling approach to knowledge building sometimes called \emph{predictive knowledge}: a collection of learning methods and architectural proposals which seek to describe the world exclusively in terms of sensation, behaviour, and time \citep{sutton_grand_2009}.

What separates predictive knowledge from other machine intelligence proposals is a focus on, and requirement of, methods which have three capacities: the ability to 1) self-verify knowledge through continual interaction with the environment, 2) describe knowledge exclusively in terms of observations from the environment, and 3) scale learning methods \citep{sutton_horde_2011}. While each of the requirements is important to operationalizing predictive knowledge, they are in service of a greater, unmentioned requirement: the ability to construct an ontology through interaction with the environment. Many other learning methods assume that ontological categories are given to the agent, either by hand-crafting properties and relationships to describe the world (as is done in knowledge bases), or providing explicit ontological categorization during supervised learning. Predictive Knowledge is liberated from this engineering process by focusing on learning methods which construct their own categories, properties, and relationships to describe the environment.


While promising, it is unclear to what extent ontologies constructed through predictive knowledge frameworks accomplish the task of conceptualizing the environment. Certainly there is some degree of success in using predictions to support control and decision-making \citep{modayil_prediction_2014,edwards_machine_2016,gunther_intelligent_2016}, but there are also roadblocks to progress which seem to indicate a problem with the foundations of the framework. For instance, tasks such as reasoning about objects in terms of sensorimotor experience in predictive terms \citep{koop_investigating_2008} has proven to be exceptionally challenging---a task which is relatively simple for supervised learning systems, and other learning methods with hand-crafted ontologies. This difficulty to express general concepts using predictions is sometimes referred to as the \emph{abstraction gap}. 


At root, and so in service of the three requirements above, the predictive knowledge project proposes that value functions hold the meaning necessary to construct a successful ontology.\footnote{This is made clear in some conceptual projects\citep{sutton_grand_2009,koop_investigating_2008,sutton_horde_2011}; however, it is worth noting that in engineering projects which focus on using predictions to support decision making, it is likely that the claim of predictions as being relevant is likely only implicit, if intended at all.} This paper will proceed by proposing a framework for understanding meaning which will present a challenge to this stated understanding of value functions, the technical implications of which will then be discussed. Importantly, we find that Predictive Knowledge does not yet meet the threshold for meaning that is necessary to accomplish the ontological construction sought after by the project of Predictive Knowledge.


\section{The Building Blocks of Meaning: a Less Deadly Triad}

There is a long standing approach to the construction of meaning in academic literature outside of Computer Science and Machine Intelligence; namely, semiotics.
Despite particular theoretical differences, the basic idea behind positions in semiotics is that meaning is a relational product of `signifiers' (or, words and language) and the `signified' (or, objects and the world). For instance, Saussure argued that the meaning for some word was defined by its relative place in the broader structure of the relevant language\textemdash i.e. \emph{dog} does not mean \emph{cat} (and vice-versa) precisely because the structure of the English language has unique places, laid out by interpreters through practice, wherein \emph{dog} and \emph{cat} cannot be interchanged. Thus, there is the signified (one's actual pet) and it stands in relation to a structured ontology of signifiers (of words one might use, or not use, to describe their pet).

Of course, put so broadly, semiotics as an approach to linguistic meaning is simultaneously compatible with a wide variety of other methodological approaches and philosophical commitments, and prone to questioning and conflict. Consequently, to see how semiotics may help design machine learning methods capable of independently constructing ontologies, the discussion must be narrowed from the broader theoretical terrain to one particular view. To that end, we focus on the Peircean model of semiotics for two reasons: 1) Peircean semiotics is not limited to the domain of language, but instead, tackles the broadest domain possible with the notion of a \emph{sign} (CP 1.339)\footnote{We refer to \emph{Collected Papers of Charles S. Peirce} by CP m.n, where m is the volume number and n is the paragraph number, as is custom in Peircean scholarship.}, making it applicable to the analysis of machine intelligence methods; and 2) Peircean semiotics particularly emphasizes the process of agent interaction rather than the resulting `language' or ontological structure (CP 1.341), making it uniquely suitable to analysis of Reinforcement Learning methods.

At a macroscopic level, Peirce's semiotics is often described as a triadic relation between an object, a signifier, and an interpretation. For example, a bonfire at a campsite could be an object; the smoke it gives off could be its signifier; and \emph{the conclusion that one draws} (or, interpretation) could be that ``people are in the forest.'' One may note that under the Peircean model, signifiers are not only words or the perspective of agents, but can also be objects in the world. This leads to two immediate confusions: 1) that there is a multiplicity to the interpretation of signs (it is not clear the the 'right' signifier is the smoke, or the fire, or that the 'right' interpretation is that there are people in the forest); and 2) that signs are not isolated or fixed, but instead are linked together (a 'complete' sign between smoke, fire and the interpretation of people, might itself be the signifier of another larger, more complex sign). However, despite these points of potential confusion, the model nonetheless provides a framework for evaluating meaning: smoke \textit{means} people are present due to the combination of relations inherent between smoke and fire, fire and people, \textit{and} the interpretive step an agent takes in relating their environment and experiences to these relations. The crux of this model is thus the third aspect of agent interpretation rather than simply the sets of relations between phenomena. 

Indeed, to go into further detail, one will find that the triadic model is derivative of Peirce's Theory of Mind.\footnote{The full scope of which is, unfortunately, beyond this paper.} What can be shown is that the semiotic model depends upon, what is normally called, the \textit{Three Categories}; and which we will refer to here as \emph{Sensation}, \emph{Perception}, and \emph{Generality}.\footnote{Peirce refers to these categories as Firstness, Secondness, and Thirdness; however, despite the systematic function that these categories play throughout his theories, we eschew this naming schema for both clarity and applicability. See CP 1.300 onwards for relevant discussion.} First, before abstractions---or conceptualizations\textemdash can be constructed, there must be the information or sensation from which one can construct the abstraction, absent of any categorization, modelling, or understanding. Thus, \emph{Sensation} is the observation an agent receives from the environment without further analysis, comparison, or relation. In the linguistic semiotic model, this would be the smoke \emph{as smoke}\textemdash a signifier without a corresponding object or interpretation, something merely sensed.

Of course, our concepts and thoughts are not simply composed of raw, unprocessed sensation. A sensation of smoke and a sensation of fire are related through our perception and environment construction \textit{regardless of any particular meaning.} Consider, for example, how classical conditioning describes fixed responses to stimuli, such as blinking, which do not require higher level conceptual cognition. Thus the second component of the triad, \emph{Perception}, describes both the properties which our environment is in terms of and how each property relates to different sensations. Within predictive knowledge and machine learning, this notion of \emph{Perception}, can be found in many places: i.e. a prediction's estimate, or the value of a state action pair.

Now, it may seem unclear why one needs a third part. There are moments sensed and perceptions which relate them; what else could there be, or must there be? Some reflection, however, may reveal the shortcomings of this diadic relationship. To think again of our campsite, one's sensation of smoke may bring about the perceptual relation of fire\textemdash the two go together after all, like blinking\textemdash but there are many further experiences which come with fire. Sometimes there really is a campsite and people roasting marshmallows, other times lightning strikes, or dry heat and unkempt brush may bring an unwelcome end to celebratory fireworks. Which conclusion one draws from the sensation of smoke\textemdash that is, what the smoke \textit{means}\textemdash is thus the application of a broader general concept (say, \textit{campfire} as compared to \textit{forest fire}). The useful selection of one broader concept over another may require a rich background of experience and learned relations (between, say, the volume of smoke and the colour of the horizon, or the smell of burning pine), but is nonetheless an active cognitive step wherein a \textit{general notion} is applied to a particular case\textemdash that the source of \textit{this} smoke, and so this fire, is people and not lightning. This third category, the relation of general patterns to particular instances, is what we have called \emph{Generality}.


\section{Is Learning a GVF a Semiotic Process?}

Thus far we have presented the parallels between AI and semiotics speculatively and hesitantly, but we will now develop an extended example covering predictive knowledge and the Three Categories. To do so, we take a General Value Function (GVF) \citep{white_developing_2015}\textemdash the most basic mechanism of many predictive knowledge proposals\textemdash and evaluate whether learning an approximate value function can be seen as a triadic relationship; we evaluate whether predicting is a process which produces meaning. 


GVFs make predictions estimating the \emph{value}, or expected discounted sum of a signal $C$ defined as $G_t = \sum^{\infty}_{k=0} (\prod^{k}_{j=1}(\gamma_{t+j}))C_{t+k+1}$. Value for some state $\phi$ is estimated with respect to a specific policy $\pi$, discount function $0 \leq \gamma \leq 1$, and cumulant $c$, such that $v(\phi; \pi, \gamma, c) = \E_\pi[G_t|S_t = \phi]$. Using GVFs, we can ask questions such as ``How long will it take me to bump into a wall if I keep walking forwards''? These GVFs are typically learnt online through interaction between an agent and its world over discrete time-steps. On each time-step $t = 0,1,2,...,n$ the agent receives a vector $o_t$ that describes what is sensed, and takes an action $a_t$. Prior to use or feature construction, the observations $o_t$ received by the system are \emph{Sensation}: the first component of the Three Categories. 

The observations, with some function approximator, are used to produce the \emph{agent state}: a feature vector $\phi_t : o_t \rightarrow \mathbb{R}^n$  which describes the environment from the agent's perspective\footnote{It is worth noting that input observations $o_t$ could include not just immediate sensor feedback, but also the previous action, historical information, or internal signals generated from learning.}. 
 This state $\phi_t$ is used in conjunction with some learning method to estimate the discounted sum $G_t$ of future signals $C$. For our example, we consider Temporal-Difference (TD) learning \citep{sutton_learning_1988}; however, our conclusions will generalize to other policy evaluation methods. When performing TD learning, we maintain some weight parameters $w \in \mathbb{R}^n$ which when combined with the current state produce the estimated return $v_\pi(\phi_t) = w_t^\top \phi_t$. On each step, at each instant, the weights are changed proportional to the TD error $\delta_t = C_{t+1} + \gamma_{t+1} v_\pi(\phi_{t+1}) - v_\pi(\phi_t)$. When the weights are updated by $w_{t+1} = w_t + \alpha_t \delta_t \phi_t$, the relation between $o_t$, $a_t$, and $c_t$ is updated based on some response from the environment. 
For any given state $\phi_t$, the estimate $v_\pi(\phi_t)$ forms a relation between what is sensed $o_t$, the actions taken $a_t$ and the signal being predicted $c_t$; thus, value estimates form the second component of the Three Categories: \emph{Perception}.





Having come to the end of the process of specifying and learning a GVF, one may wonder where \emph{Generality} exists in predictive knowledge. After all, many claim that a single prediction has meaning\footnote{See \cite{sutton_horde_2011} for an example argument for GVFs as having explicit semantics, and both \cite{koop_investigating_2008} and \cite{white_developing_2015} for additional discussion of predictions as inherently meaningful.}, but we have so far only identified \emph{Sensation} and \emph{Perception}. While a GVF may capture a prediction for any given state, \emph{generalizing} over observations through some function approximation, it does not capture \emph{Generality}. When a prediction is formed as a GVF, our expectation of future signals slurs over all experience, making it impossible to relate manifestations of instances of signals in order to compare and contrast them. Using GVFs alone, we are incapable of, say, identifying that a wall bumped into is the very same as the one we bumped into both 10 time-steps ago and 100 time-steps ago: we may only say how close an observation was to the expectation of the whole of an agent's experience in that particular agent-state.


This limitation in expressing the Three Categories invites us to wonder whether the notion of Sensation, Perception, and Generality is a productive one: does framing predictive knowledge as a semiotic process help us better understand machine intelligence? As we previously introduced, there are other approaches to semiosis, some of which do not depend on \emph{Generality}\footnote{For example, \citep{barbieri_introduction_2007} presents a variety of semiotic models in application to cell biology.}. Simply finding our methods to be meaningful does not obliviate the limitations of existing predictive knowledge methods. Predictive knowledge frameworks can be construed as constructing meaning under other definitions of semiotics; however, declaring our methods to be sufficient does not help predictive knowledge systems cross the abstraction gap and express concepts which are at present elusive---a declaration of meaning would not suddenly enable Predictive Knowledge methods to reason about generality, or make it any clearer how notions such as objects would be formalized in a predictive setting.

How, then, do we surmount the gap between abstract generalities and the relations which inform them? While predictions alone are insufficient, it may be possible to express generalities by constructing models using predictions
A model-based method which explicitly defines state-action transitions relates not only one state $s_t$ to another $s_{t+1}$, but relates states in terms of all the possible transitions given all the possible actions which could have been taken in $s_{t}$. A model which is able to interrelate many predictions such that contexts can be compared and contrasted could be considered semiotic. For these reasons, difficulty in constructing generalities does not belong to insufficient learning methods, or poor state construction (although they do impact progress). The difficulty of constructing abstractions results from an inability to interrelate what is learned: it is a problem of how we structure predictive knowledge architectures, not how we learn them.

Does learning a prediction encompass the entirety of a semiotic process? No; however, predictive knowledge could play a central role in a process which is semiotic. GVFs provide a robust and flexible foundation for semiosis by playing the part of \emph{Perception}. By focusing on incremental learning methods which are specified in terms of behaviour, GVFs describe a class of predictions which are uniquely suited to be Perception: methods which are not ontologically constrained by labels\textemdash or,
what predictive knowledge agents can conceptualize constrained by the reality of the environment they inhabit, not the labels provided for training. Moreover, GVFs have proven themselves to be practically useful for reactive behaviour such as prosthetic control \citep{edwards_machine_2016}, lazer welding \citep{gunther_intelligent_2016}, and robotics \citep{modayil_prediction_2014}---crucial steps in demonstrating that GVFs are useful in informing decision-making about the environment, although insufficient for constructing an ontology of the world.


\section{Conclusion: Taking Stock of the Predictive Knowledge Project}

Predictive knowledge describes a collection of proposals for constructing machine knowledge which assert that all world knowledge can be described as predictions about sensation, behaviour, and time. In this paper, we take a first look at the construction of semantics and meaning in predictive knowledge by evaluating whether or not learning a General Value Function can be construed as a semiotic process. We demonstrate that GVFs can be seen to fulfill the first two components of semiosis: \emph{Sensation} and \emph{Peception}; however, predictions fall short of providing the third component, \emph{Generality}. As a result, predictions do not have meaning independent of any other process. While predictive knowledge proposals do not presently construct meaning, the project of predictive knowledge is not inherently doomed; quite the opposite, predictive knowledge provides a promising foundation for construction of meaning in Machine Intelligence. We suggest that it may be possible to express generalities by using predictions to construct models of the environment, thereby completing the triadic relation and forming a semiotic process.

%
%

\vspace{-1em}
\nocite{peirce_collected_1931}
\bibliography{My_Library.bib}

\end{document}